\documentclass[10pt,twocolumn,letterpaper]{article}

\usepackage[pagenumbers]{cvpr} % To force page numbers, e.g. for an arXiv version

\usepackage[dvipsnames]{xcolor}
\usepackage{pifont}

\usepackage[table]{xcolor}
\definecolor{tablered}{HTML}{FDE0E0}
\definecolor{tableorange}{HTML}{FFE6CC}
\definecolor{tableyellow}{HTML}{FFF4CC}
\usepackage{graphicx}
\usepackage{epigraph,varwidth}
\usepackage{booktabs}   % For \toprule, \midrule, \bottomrule
\usepackage{multirow}   % For multirow functionality
\usepackage{array}      % For better table formatting
\usepackage{colortbl} % For row colors
\usepackage{tabularx} % For tabularx environment
\usepackage{graphicx} % For table adjustments
\usepackage{booktabs} % For professional-looking tables
\usepackage{caption} % For table captions

\usepackage{xcolor,graphicx}

\definecolor{cvprblue}{rgb}{0.21,0.49,0.74}
\usepackage[pagebackref,breaklinks,colorlinks,citecolor=cvprblue]{hyperref}

\def\paperID{} % *** Enter the Paper ID here
\def\confName{CVPR}
\def\confYear{2026}

\title{WildRayZer: Self-supervised Large View Synthesis in Dynamic Environments}

\author{
Xuweiyi Chen, \space Wentao Zhou, \space Zezhou Cheng \\
University of Virginia \\ 
[0.5em]
\texttt{\url{https://wild-rayzer.cs.virginia.edu/}}
}

\begin{document}
\maketitle
\begin{abstract}
We present \textbf{WildRayZer}, a self-supervised framework for novel view synthesis (NVS) in dynamic environments, where both the camera and objects move. Dynamic content breaks the multi-view consistency that static NVS models rely on, causing ghosting, hallucinated geometry, and unstable pose estimation. WildRayZer addresses this by performing an analysis-by-synthesis test: a camera-only static renderer explains rigid structure, and its residuals reveal transient regions. From these residuals, we construct pseudo motion masks, distill a motion estimator, and use it to mask input tokens and gate loss gradients so supervision focuses on cross-view background completion. To enable large-scale training and evaluation, we curate Dynamic RealEstate10K (D-RE10K), a real-world dataset of 15K casually captured dynamic sequences, and D-RE10K-iPhone, a paired transient and clean benchmark for sparse-view transient-aware NVS. Experiments show that WildRayZer consistently outperforms optimization-based and feed-forward baselines in both transient-region removal and full-frame NVS quality with a single feed-forward pass.

\vspace{-10pt}
\end{abstract} 
\section{Introduction}
\label{sec:intro}

Self-supervised learning from large-scale unlabeled data with minimal handcrafted inductive bias has driven major advances in LLMs~\cite{radford2019language,kaplan2020scaling} and visual understanding~\cite{he2022masked,oquab2023dinov2} and generation~\cite{peebles2023scalable}. 
A similar trend is only beginning to emerge in 3D vision. 
While recent 3D models achieve strong performance~\cite{hong2024lrm,wang2024dust3r,wang2025vggt}, they often rely on accurate 3D annotations such as point maps and camera poses, or on explicit 3D inductive biases in representations such as NeRF~\cite{mildenhall2020nerf} and 3D Gaussian Splatting~\cite{kerbl20233d}. These requirements limit scalability, robustness in dynamic environments, and out-of-distribution generalization.

\begin{figure}[t]
    \centering
    \includegraphics[width=\linewidth]{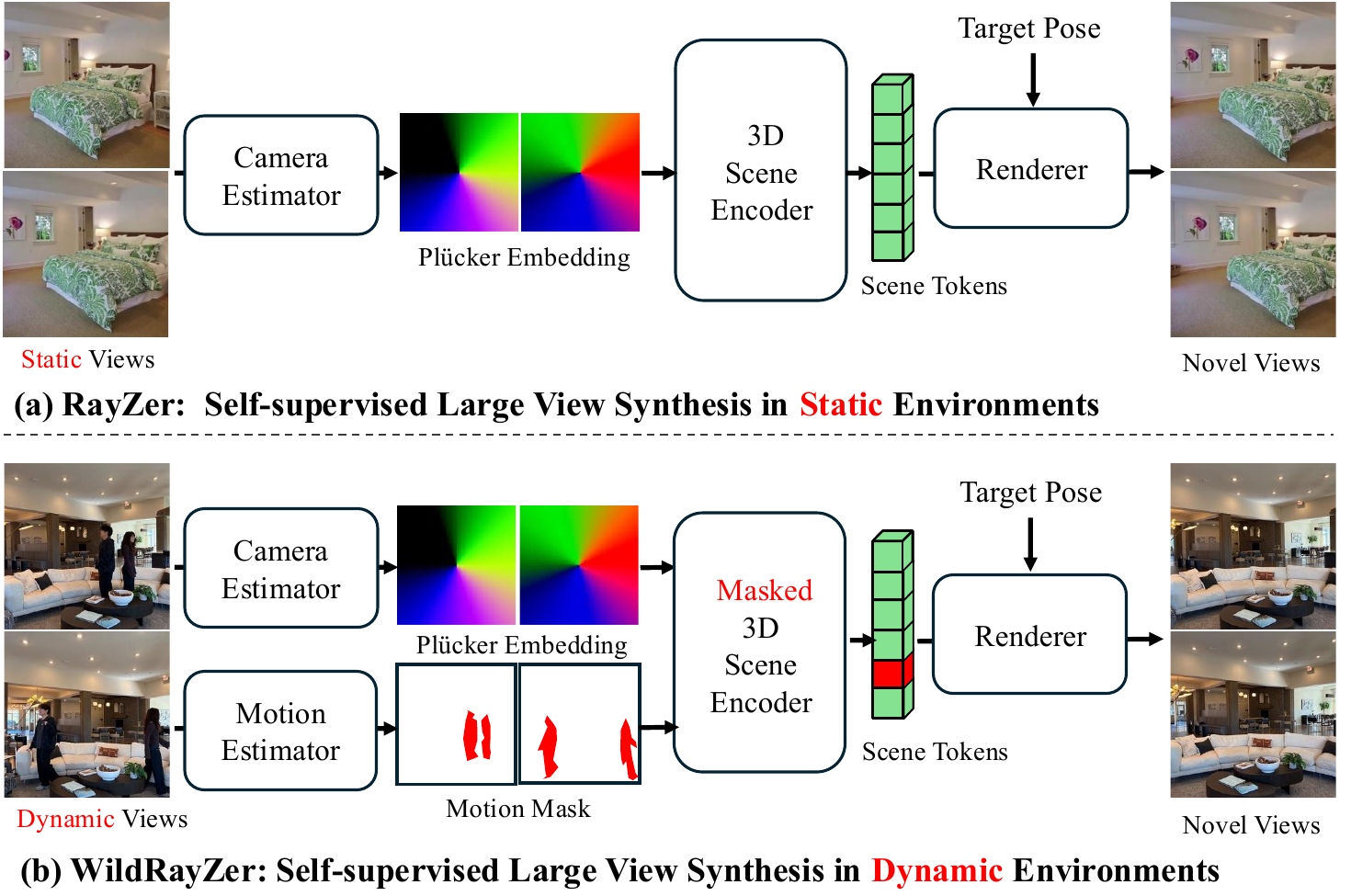}
    \vspace{-7mm}
    \caption{
    Our self-supervised \textbf{WildRayZer} learns to render static novel views from \emph{dynamic} images without any 3D or GT mask supervision. It extends the state-of-the-art self-supervised large view synthesis model RayZer to dynamic environments by adding a learned motion mask estimator and a masked 3D scene encoder.
    }
    \label{fig:teaser}
    \vspace{-6mm}
\end{figure}
Motivated by this, recent work has explored reducing or removing explicit 3D supervision or inductive bias. 
For example, LVSM~\cite{jin2024lvsm} proposes a transformer-based view synthesis model with minimal 3D inductive bias.
NoPoSplat~\cite{ye2024no} eliminates the need for camera pose at inference; SPFSplat~\cite{huang2025no} and Splatt3R~\cite{smart2024splatt3r} removes pose supervision during training.
Building on LVSM, RayZer~\cite{jiang2025rayzer} introduces a self-supervised framework for sparse-view novel view synthesis that does not require any 3D supervision.

Despite recent progress~\cite{mitchel2025true,zhao2025rayzer}, prior work still relies on a fundamental assumption: \emph{\textbf{the 3D scene is static}}, as illustrated in \Cref{fig:teaser}(a).
These models require static inputs during both training and inference, yet real-world 3D environments are inherently dynamic. 
As a result, existing methods rely on static scene datasets such as RealEstate10K~\cite{zhou2018stereo}, which restricts scalability and prevents full use of abundant in the wild videos with natural dynamic content.

In this work, we propose \textbf{WildRayZer}, a self-supervised learning framework for novel view synthesis in dynamic environments from sparse, unposed views. 
Similar to prior efforts that adapt NeRF and 3DGS to in-the-wild settings~\cite{ren2024nerf,martin2021nerf,xu2024wild,zheng2025wildgs,kulhanek2024wildgaussians}, WildRayZer extends RayZer to dynamic scenes while removing dynamic objects in the final renderings.
As in RayZer, WildRayZer takes unposed multiview images as input, reconstructs the 3D scene implicitly, and renders novel views (See \Cref{fig:teaser}(a)); Differently, our inputs include both camera motion and object dynamics (See \Cref{fig:teaser}(b)).
Achieving this capability requires solving two key challenges.

\emph{\textbf{First, how can we localize dynamic objects without any groundtruth dynamic mask supervision?}}
WildRayZer adopts an analysis-by-synthesis strategy: a static renderer predicts what a rigid scene should look like, and deviations from this prediction identify dynamic regions. 
Specifically, we begin with a pretrained RayZer and derive pseudo motion masks from its rendering error, computed using a combination of DINOv3~\cite{simeoni2025dinov3} features and SSIM~\cite{wang2004image}. 
A motion mask head is then distilled from these masks. 
Inspired by prior work in instance segmentation~\cite{wang2024videocutler,ghiasi2021simple}, 
we further improve the robustness of this module using a simple copy paste augmentation strategy to enrich dynamic mask supervision with synthetic examples. 

\emph{\textbf{Second, how can we train and evaluate our model on this task when existing large-scale 3D datasets capture mainly static scenes?}}
Common real-world datasets for novel view synthesis (Table~\ref{tab:dataset_comparison}) contain only static scenes and therefore cannot support systematic training or evaluation of methods that separate static structure from dynamic objects. 
To enable controlled studies of this setting,
we construct Dynamic RealEstate-10K, as a natural extension of RealEstate-10K, which contains over 15K casually captured indoor sequences with moving cameras and moving objects such as humans and pets. 
We also collect a benchmark with paired transient and clean views of the same scenes for evaluation. 
Experiments and qualitative results on both existing datasets and our newly collected benchmarks show that WildRayZer outperforms prior baselines in novel view synthesis and dynamic motion prediction in a large margin.

Our main contributions are summarized as follows:
\begin{itemize}[itemsep=0pt, topsep=1pt, leftmargin=10pt]
\item We propose WildRayZer, a self-supervised learning framework for novel view synthesis in dynamic environments without any supervision on camera poses or dynamic region masks, with sparse unposed views as input.
\item We collect Dynamic RealEstate-10K, a large-scale video dataset captured in dynamic scenes, as a natural extension of RealEstate-10K, complementary to the commonly used real-world video datasets for 3D vision tasks.   
\item Our experiments show the superior performance of the WildRayZer in both NVS from sparse dynamic inputs and the motion segmentation tasks. 
\end{itemize}

\section{Related Work}
\label{sec:related}

\paragraph{Optimization-based Novel View Synthesis.}
NeRF~\cite{mildenhall2020nerf} introduced a neural volumetric 3D representation with differentiable volume rendering, enabling neural scene reconstruction by minimizing a photometric rendering loss and achieving remarkable novel view synthesis quality. However, classical NeRF pipelines critically rely on clean camera poses and static scenes, which severely limits their applicability in realistic, in-the-wild settings. This has motivated a line of work on robust NVS that relaxes these assumptions and seeks to make NeRF viable in the wild~\cite{martin2021nerf,rematas2022urban,tancik2022block,sabour2023robustnerf,chen2024nerf,schonberger2016structure}. Beyond these, explicit warping pipelines~\cite{yoon2020novel} and early pose-free NVS~\cite{liu2020auto3d} offer alternatives, but can be brittle under in-the-wild dynamics settings.

More recently, 3D Gaussian Splatting further improves rendering quality and speed by representing scenes with explicit 3D Gaussians and fast rasterization, but inherits the same reliance on accurate poses and static scenes. Another line of work trains Gaussian splatting in the wild by modeling appearance variation with learned appearance embeddings, per-primitive color adaptation, or spatial appearance fields~\cite{zhang2024gaussian,dahmani2024swag,wang2024we,xu2024wild}, typically down-weighting high-error pixels following the robust NeRF line of work~\cite{martin2021nerf}. In contrast, we perform NVS with a large transformer-based renderer without specific 3D representation such as NeRF or 3DGS, and explicitly target dynamic, in-the-wild scenarios.

\vspace{-5mm}
\paragraph{Generalizable Novel View Synthesis.}
Generalizable methods enable fast NVS by training neural networks across many scenes to predict novel views or an underlying 3D representation in a single forward pass. Early work such as PixelNeRF~\cite{yu2021pixelnerf}, MVSNeRF~\cite{chen2021mvsnerf}, and IBRNet~\cite{wang2021ibrnet} predicts volumetric 3D features from input views, leveraging strong 3D inductive biases like epipolar geometry and plane-sweep cost volumes. Subsequent methods improve robustness under sparse views~\cite{liu2022neural,johari2022geonerf,jiang2024few,szymanowicz2024splatter,jiang2023leap}, while others extend these ideas to 3DGS-based scene representations~\cite{charatan2024pixelsplat,szymanowicz2025flash3d,chen2024mvsplat,tang2024lgm}.

More recently, 3D large reconstruction models (LRMs)~\cite{hong2024lrm,xie2024lrm,wang2024pflrm,zhang2024gslrm,ziwen2024long,ma20254d} adopt scalable transformer architectures trained on large-scale data to learn generic 3D priors. LVSM~\cite{jin2024lvsm} goes a step further by largely removing hand-crafted 3D inductive biases and learning a powerful token-space renderer, leading to improved fidelity and scalability—but still assumes known poses and static imagery. RayZer~\cite{jiang2025rayzer} follows a similar transformer-based direction and introduces a self-supervised, pose-free framework for sparse-view NVS, yet remains restricted to \emph{static} scenes. Our method builds on this line of work and explicitly lifts the static-scene assumption, targeting dynamic, in-the-wild inputs where both cameras and objects move.

\vspace{-5mm}
\paragraph{Moving Object Segmentation.}
Most moving-object segmentation approaches are explicitly video-based and fall into three families. (a) \emph{Flow-based} methods~\cite{bideau2016s,papazoglou2013fast,sekkati2007variational,wedel2009detection} detect motion by grouping optical-flow cues; (b) \emph{Trajectory-based} approaches~\cite{arrigoni2020usage,jiang2021select,xu2018motion,homeyer2023moving,neoral2021monocular,huang2024zero,huang2025segment} reason over multi-frame geometric or point trajectories; (c) \emph{Unsupervised Video Object Segmentation} (UVOS)~\cite{perazzi2016benchmark,pont20172017,yang2021learning,zhou2020matnet,ren2021reciprocal,ji2021full,wang2024videocutler,liu2021emergence} typically segments salient objects, and thus can capture static entities rather than true movers. Instead of operating on explicit video sequences and point trajectories, we derive motion evidence from what a static multi-view renderer \emph{cannot} explain. This learned mask predictions become reliable under sparse-view unposed setting.
\section{Dynamic RealEstate10K}
\label{sec:dre10k}

We target novel view synthesis from dynamic, in-the-wild videos where both the camera and scene undergo motion. 
Rather than relying on controlled captures, we mine diverse handheld footage from public sources. 
A key observation is that user-generated real-estate walkthroughs naturally provide abundant multi-view coverage of cluttered indoor scenes. 
Unlike prior static NVS datasets, which aggressively filter out motion, we intentionally \emph{retain} clips containing humans, pets, and object interactions to expose transient dynamics. 
As shown in Table~\ref{tab:dataset_comparison}, static NVS datasets are large, but existing dynamic datasets remain extremely limited in size, often containing less then ten sequences due to their optimization-heavy pipelines. 
To close this gap, we collect a large-scale dynamic scene video dataset, namely Dynamic RealEstate10K (D-RE10K), which consists of 15K real indoor sequences. 
Following standard practice for web-mined datasets (\eg, WebVid-2M~\cite{Bain21}, RealEstate10K~\cite{zhou2018stereo}),  we will release annotations, metadata, and video links under a CC license and we have anonymized faces using EgoBlur~\cite{raina2023egoblur}.

\subsection{Data Curation}

Our curation pipeline proceeds in three stages:

\noindent
\emph{(1) Source identification}.
We start by querying YouTube channels that primarily publish real-estate walkthroughs and indoor pet-interaction videos. These videos provide smooth handheld motion, stable lighting, and cluttered but mostly rigid backgrounds, which is ideal for multi-view reconstruction. 
We filter based on metadata to exclude overly short or low-quality uploads.

\noindent
\emph{(2) Image-level filtering}. 
From each candidate clip, we uniformly sample frames and assess visual quality using image quality assessment~\cite{su2020blindly}, discarding low-quality or heavily compressed videos. 
We further apply an OCR-based text detector~\cite{baek2019character} to remove videos containing intrusive overlays (\eg, subtitles, watermarks, real-estate banners).

\noindent
\emph{(3) Sequence extraction}.
NVS requires geometric coherence within a localized region. We apply TransNet V2~\cite{soucek2024transnet} to detect scene cuts and discard transitions. Within each shot, we estimate camera trajectories using DPVO~\cite{teed2023deep} and subdivide clips based on translation magnitude to ensure sufficient parallax and consistent motion. The resulting sequences exhibit stable camera motion and meaningful dynamic content.

\subsection{Benchmark Construction}
We build two evaluation splits: (1) the \emph{D-RE10K motion-mask benchmark}, providing motion annotations for Internet videos, and (2) \emph{D-RE10K-iPhone}, a real-world paired transient/clean dataset for sparse-view transient-aware NVS.

\newcommand{\cmark}{\textcolor{green!60!black}{\ding{51}}} % green check
\newcommand{\xmark}{\textcolor{red!70!black}{\ding{55}}}   % red cross

\begin{table}[t]
\centering
\scriptsize
\setlength{\tabcolsep}{5pt}
\begin{tabular}{lccccc}
\toprule
\textbf{Dataset} & \textbf{\#Seq.} & \textbf{Real?} & \textbf{Dynamic?} & \textbf{Large?} & \textbf{Scene Type} \\
\midrule
DL3DV~\cite{ling2024dl3dv}                 & 10K     & \cmark & \xmark & \cmark & Mixed \\
RealEstate10K~\cite{zhou2018stereo}        & 80K     & \cmark & \xmark & \cmark & Indoor \\
CO3D~\cite{reizenstein2021common}          & 19K     & \cmark & \xmark & \cmark & Object \\
Matterport3D~\cite{chang2017matterport3d}  & 10K     & \cmark & \xmark & \cmark & Indoor \\
Objaverse-XL~\cite{deitke2023objaverse}    & 10M     & \xmark & \xmark & \cmark & Object \\
\midrule
Plenoptic Video~\cite{li2022neural}        & 6       & \cmark & \cmark & \xmark & Indoor \\
D-NeRF~\cite{pumarola2021d}                & 8       & \xmark & \cmark & \xmark & Object \\
NSFF~\cite{li2021neural}                   & 8       & \cmark & \cmark & \xmark & Mixed \\
Nerfies~\cite{park2021nerfies}             & 9       & \cmark & \cmark & \xmark & Object \\
HyperNeRF~\cite{park2021hypernerf}         & 17      & \cmark & \cmark & \xmark & Object \\
DyCheck~\cite{gao2022monocular}            & 14      & \cmark & \cmark & \xmark & Object \\
RobustNeRF~\cite{sabour2023robustnerf}     & 5       & Mixed  & \cmark & \xmark & Object \\
NeRF-On-the-Go~\cite{ren2024nerf}          & 12      & \cmark & \cmark & \xmark & Mixed \\
WildGS-SLAM~\cite{zheng2025wildgs}         & 10      & \cmark & \cmark & \xmark & Indoor \\
T-3DGS~\cite{markin2024t}                  & 5       & \cmark & \cmark & \xmark & Indoor \\
\midrule
\textbf{D-RE10K (Ours)}                    & \textbf{15K} & \cmark & \cmark & \textbf{\cmark} & Indoor \\
\bottomrule
\end{tabular}
\vspace{-1mm}
\caption{\textbf{Common datasets for novel view synthesis.}
“Large?” marks datasets with $\geq$10K sequences. 
Static NVS datasets are large, but dynamic ones are typically tiny. 
\textbf{D-RE10K} closes this gap with \emph{15K} real, in-the-wild and dynamic indoor sequences featuring diverse transient objects (people, pets, clutter), enabling training for transient-aware NVS at scale.\vspace{-2mm}  }
\label{tab:dataset_comparison}
\end{table}
\noindent
\textbf{D-RE10K Motion Mask.}
To evaluate transient-region estimation performance on Internet videos, we derive per-frame motion masks by fusing coarse cues from MegaSAM~\cite{li2025megasam} with image segmentation from SAM2~\cite{ravi2024sam}. The fused masks are human-verified for accuracy. 
We curate 25 sequences for validation and 74 for testing. On the test split, we use ground-truth to analysis mask quality.  Moreover, we report masked-PSNR, masked-SSIM, and masked-LPIPS~\cite{zhang2018perceptual} restricted to static regions in order to ensure fair comparison of NVS quality.

\noindent
\textbf{D-RE10K-iPhone.}
Although dynamic NVS datasets exist, such as NeRF-On-the-Go~\cite{ren2024nerf} and WildGS-SLAM~\cite{zheng2025wildgs}, their train and test splits are often captured under distinct camera trajectories and viewpoint.
In our sparse-view regime with only \emph{2–4 input views}, reliable evaluation requires substantial spatial overlap between inputs and targets; otherwise, reconstruction errors become entangled with coverage gaps rather than modeling quality. Moreover, datasets like RobustNeRF~\cite{sabour2023robustnerf} contain only two highly controlled tabletop scenes, making it difficult to draw conclusions about performance in real, in-the-wild environments.

To address this, we construct D-RE10K-iPhone, a 50-sequence benchmark for real-world, sparse-view transient-aware NVS (\ie, handling transient objects). Using a tripod-mounted iPhone, we reposition the camera between views to ensure geometric diversity and, for each viewpoint, capture a pair of images: one containing a transient object or multiple transient objects (\eg, person, vehicle) and one without, triggered via Bluetooth to minimize pose drift. The resulting matched pairs span kitchens, studies, dining rooms, and living spaces with paintings and decor, enabling fine-grained evaluation of transient-region reconstruction and full-frame fidelity under realistic conditions. Additional details are provided in the appendix.

\begin{figure*}[t]
    \centering
    \includegraphics[width=\linewidth]{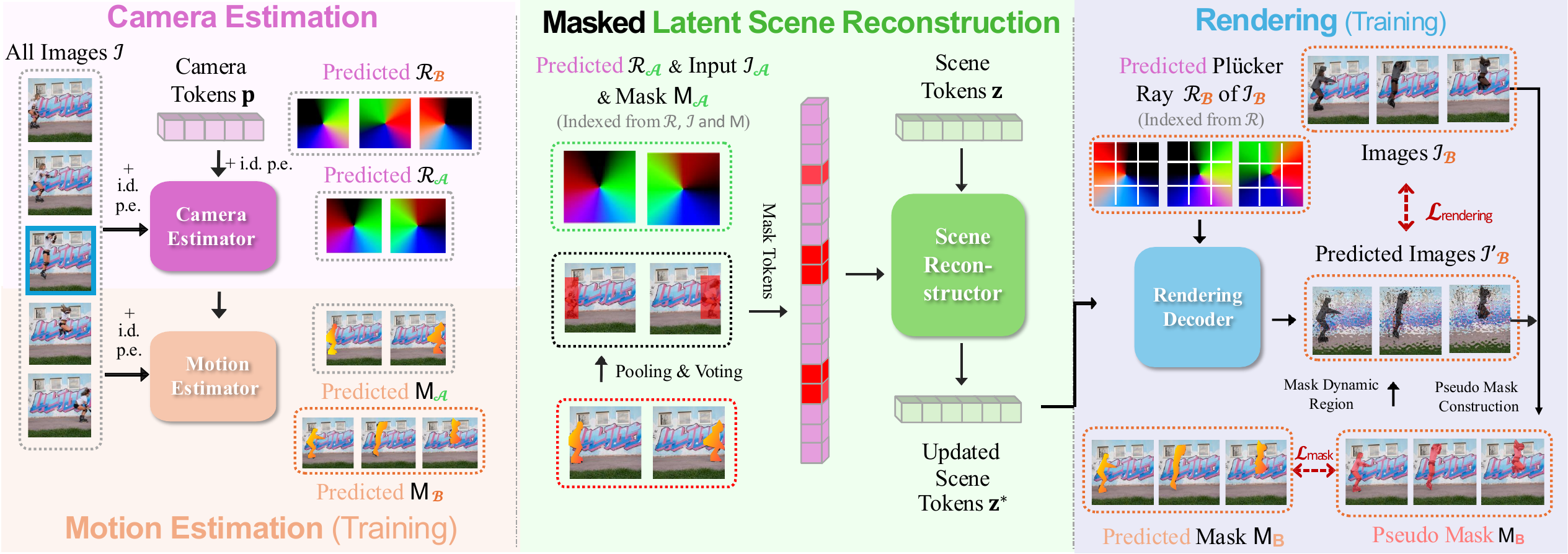}
    \vspace{-4mm}
    \caption{
        \textbf{WildRayZer self-supervised learning framework.}
        \textbf{(a) Training.} WildRayZer takes unposed, uncalibrated multi-view \emph{dynamic} images $\mathcal{I}$ and predicts per-view camera parameters (intrinsics and relative poses), which are converted into pixel-aligned Plücker ray maps $\mathcal{R}$. A camera-only static renderer explains the rigid background; residuals between renderings $\hat{\mathcal{I}}_B$ and targets $\mathcal{I}_B$ highlight dynamic regions, which are sharpened by our pseudo-motion mask constructor (see \Cref{subsec:psuedo} and \Cref{fig:pseudomask}). We distill a motion estimator from these pseudo-masks and use it to gate dynamic image tokens before scene encoding; the same pseudo-masks also gate dynamic pixels in the photometric rendering loss.
        \textbf{(b) Inference.} Given dynamic input views $\mathcal{I}$, the model predicts camera parameters, motion masks, and a static scene representation in a single feed-forward pass. The motion estimator operates on the input views to mask dynamic tokens, and the renderer synthesizes transient-free novel views given the inferred scene representation and a target camera.
    }
    \label{fig:main}
    \vspace{-2mm}
\end{figure*}
\section{Methodology}

In this section we detail the WildRayZer training pipeline. We first review RayZer in \Cref{subsec:pre-rayzer}, then present the details of the WildRayZer architecture and learning scheme in \Cref{subsec:wildrayzer}.

\subsection{Preliminaries}
\label{subsec:pre-rayzer}
RayZer~\cite{jiang2025rayzer} is a self-supervised, transformer-based \emph{token-space renderer} for novel view synthesis in \emph{static} scenes, trained without any 3D supervision (no ground-truth geometry or camera poses). The \textbf{input} is a set of static, unposed, uncalibrated images
\(\mathcal{I}=\{I_i\in\mathbb{R}^{H\times W\times3}\mid i=1,\dots,K\}\),
where \(H\times W\) is the image resolution and \(K\) is the number of input views.
Each \(I_i\) is split into non-overlapping \(s{\times}s\) patches (patch size \(s\)), and linearly embedded following ViT~\cite{dosovitskiy2020image} into tokens
\(f_i\in\mathbb{R}^{h\times w\times d}\),
where \(h=H/s\), \(w=W/s\) denote the patch-grid height and width, and \(d\) is the token embedding dimension. RayZer uses the sum of sinusoidal \emph{spatial} and shared \emph{image-index} positional encoding~\cite{dosovitskiy2020image,bertasius2021space}.

A camera estimator \(\mathcal{E}_{\mathrm{cam}}\) predicts per-view rigid poses
\(\mathbf{P}_i\in\mathrm{SE}(3)\) and shared intrinsics \(\mathbf{K}\).
RayZer splits \(\mathcal{I}\) into two disjoint subsets, \(\mathcal{I}_{\mathcal{A}}\) (inputs) and \(\mathcal{I}_{\mathcal{B}}\) (targets). The scene reconstructor \(\mathcal{E}_{\mathrm{encode}}\) encodes \(\mathcal{I}_{\mathcal{A}}\) into a scene representation
\(z\in\mathbb{R}^{L\times d}\)
with \(L\) scene tokens, conditioned on
\(\mathcal{P}_{\mathcal{A}}=\{(\mathbf{P}_i,\mathbf{K})\mid I_i\in\mathcal{I}_{\mathcal{A}}\}\).
A rendering decoder \(\mathcal{D}_{\mathrm{render}}\) then synthesizes the held-out views using a transformer decoder~\cite{jin2024lvsm}. Given \(z\) and tokenized Plücker rays \(r\) derived from
\(\mathcal{P}_{\mathcal{B}}=\{(\mathbf{P}_i,\mathbf{K})\mid I_i\in\mathcal{I}_{\mathcal{B}}\}\),
the decoder produces rendered targets \(\hat{\mathcal{I}}_{\mathcal{B}}\).
The self-supervised objective compares rendered and ground-truth targets:
\begin{align}
\label{eq:loss}
\mathcal{L}
= \frac{1}{|\mathcal{I}_{\mathcal{B}}|}\!\sum_{\hat{I}\in\hat{\mathcal{I}}_{\mathcal{B}}}
\Big(\mathrm{MSE}(I,\hat{I})+\lambda\,\mathrm{Percep}(I,\hat{I})\Big),
\end{align}
where each \(\hat{I}\) is paired with its ground-truth \(I\in\mathcal{I}_{\mathcal{B}}\), and \(\mathrm{Percep}(\cdot,\cdot)\) is a perceptual loss.

\subsection{WildRayZer}
\label{subsec:wildrayzer}

\paragraph{Overview.}
RayZer recovers cameras and a scene representation from unposed, uncalibrated inputs but assumes static imagery. WildRayZer lifts this restriction: given dynamic, in-the-wild inputs, it aims to render the underlying \emph{static} 3D scene by disentangling transient object motion from camera motion in a fully self-supervised manner. Concretely, we augment RayZer with a motion estimator $\mathcal{E}_{\mathrm{mot}}$ placed alongside the camera estimator $\mathcal{E}_{\mathrm{cam}}$, scene encoder $\mathcal{E}_{\mathrm{scene}}$, and renderer $\mathcal{D}_{\mathrm{render}}$. To prevent dynamic content from contaminating the static scene tokens, we adopt an alternating optimization schedule: first learn motion masks while freezing the renderer stack, and then learn a masked renderer while freezing the motion head. Once the masks are sufficiently reliable, we jointly fine-tune all components. Throughout training, there is no supervision from ground-truth poses, depth, or semantic labels; all pseudo-labels are derived from discrepancies between rendered and observed images, and external backbones (\eg, DINOv3) remain frozen.

To illustrate the pipeline, we show the learning framework in \Cref{fig:main}.
When training the motion estimator, $\mathcal{E}_{\mathrm{cam}}$, $\mathcal{E}_{\mathrm{scene}}$, and $\mathcal{D}_{\mathrm{render}}$ are fixed. For each held-out view $I\in\mathcal{I}_{\mathcal{B}}$, the static renderer predicts a target $\hat I$, and we construct a soft pseudo motion mask $\tilde M(I)\in[0,1]^{H\times W}$ by treating appearance and feature-space residuals between $I$ and $\hat I$ as evidence of dynamics. The motion head $\mathcal{E}_{\mathrm{mot}}$ predicts per-pixel logits $S(I)\in\mathbb{R}^{H\times W}$, which we train to match $\tilde M(I)$ using a standard BCE-with-logits objective. In the complementary phase, when training the renderer, the motion estimator is frozen. For each input view $I\in\mathcal{I}_{\mathcal{A}}$, we convert $S(I)$ to probabilities, downsample to the patch grid, threshold to obtain a binary patch mask $\Pi\in\{0,1\}^{h\times w}$, and zero out dynamic token positions in the fused token map before encoding with $\mathcal{E}_{\mathrm{scene}}$. This yields a static scene representation $z$ that is explicitly purged of transient content, and we then optimize the reconstruction loss in Eq.~\eqref{eq:loss} on the held-out targets $\mathcal{I}_{\mathcal{B}}$. Finally, to improve robustness to open-set distractors, we apply a simple copy–paste augmentation~\cite{ghiasi2021simple, wang2024videocutler}: COCO~\cite{lin2014microsoft} objects with ground-truth masks are pasted into training views and their masks are treated as additional $\tilde M(I)$, providing precise synthetic transients without altering cameras or the rendering architecture. At this stage, we train the model end to end with augmented data and dynamic data, as illustrated in \Cref{fig:main}.

\paragraph{Motion Estimator.}
The motion estimator $\mathcal{E}_{\mathrm{mot}}$ predicts, for each image $I$, a per-pixel logit map $S(I)\in\mathbb{R}^{H\times W}$. To obtain a robust motion probability map under noisy pseudo labels, we fuse three complementary signals that are all aligned on the $h\times w$ token grid: (a) DINOv3~\cite{simeoni2025dinov3} patch features, (b) RayZer image tokens, and (c) Pl\"{u}cker–ray tokens derived from $(\mathbf{P},\mathbf{K})$. For each token position $(i,j)$, we take the corresponding DINOv3 feature, image token, and ray token, apply LayerNorm and a learned linear projection to a shared width $d$, and concatenate them. A small fusion MLP then produces a unified token representation that serves as input to a shallow transformer stack.

Tokens from all input views are concatenated along the sequence dimension to enable cross-view reasoning, and a DPT-style~\cite{Ranftl2020} decoder upsamples the fused tokens back to $H\times W$ to produce $S(I)$ for each image. We find that including DINOv3 features accelerates convergence and yields sharper, more semantically aligned motion masks. Importantly, for the image tokens we use the features \emph{before} the camera estimator $\mathcal{E}_{\mathrm{cam}}$ so that the motion head never relies on target views or test-time-only signals: at inference, it sees exactly the same type of inputs as during training. The motion estimator is then trained with a standard BCE-with-logits loss to match the pseudo labels $\tilde M(I)$.

\begin{figure}[t]
\centering
\includegraphics[width=\linewidth]{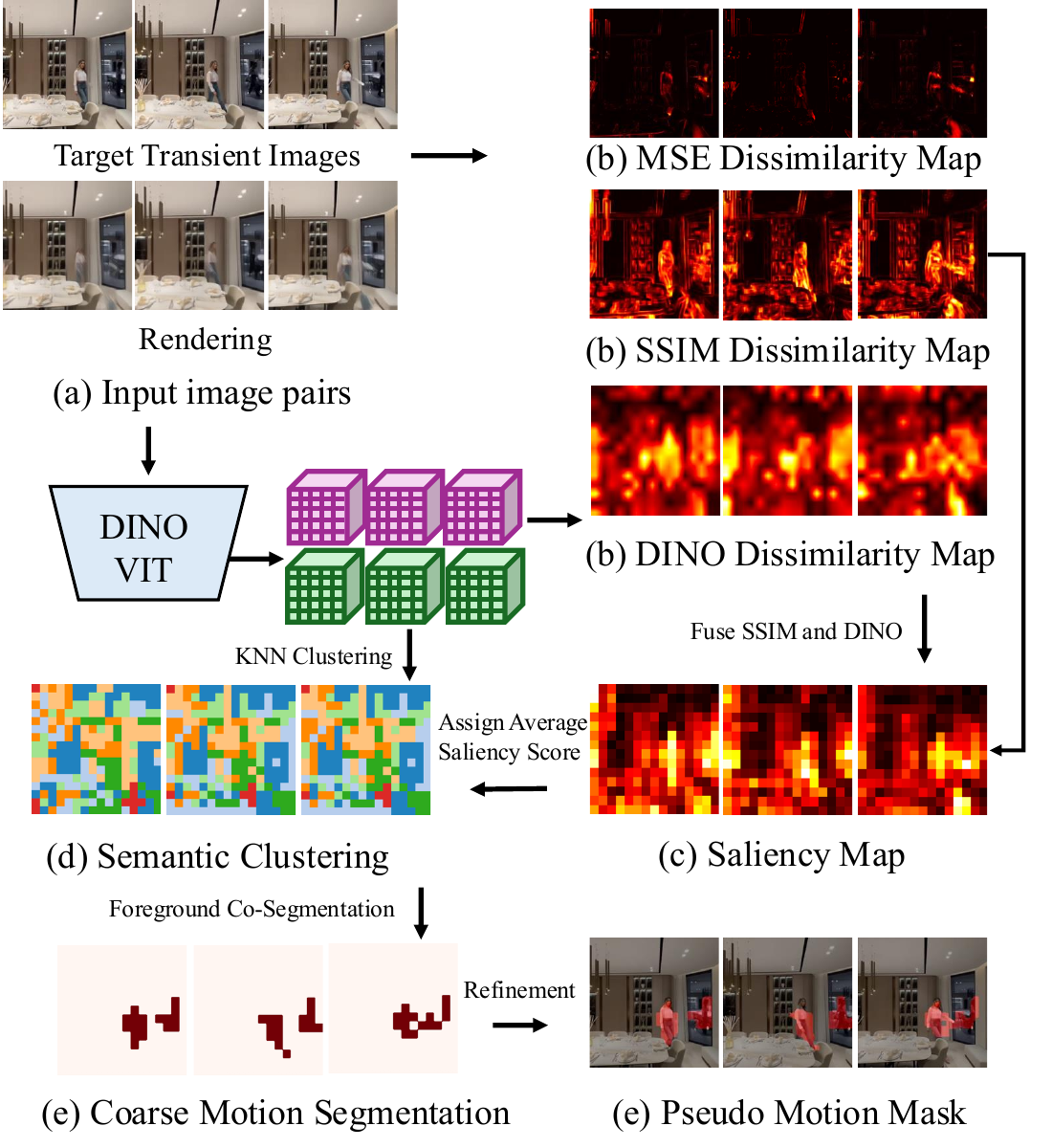}
\vspace{-4mm}
\caption{\textbf{Pseudo Motion Mask Pipeline.} We fuse SSIM- and DINO-based dissimilarity into a saliency map, cluster DINO patch features to vote for dynamic patches, then refine the coarse patch mask to pixel resolution via morphological smoothing, small-component removal, and GrabCut~\cite{rother2004grabcut}.\vspace{-4mm}}
\label{fig:pseudomask}
\end{figure}
\paragraph{Pseudo-label Construction.}
\label{subsec:psuedo}

Given a target image $I$ and its static rendering $\hat{I}$, we construct motion pseudo-labels by fusing semantic and appearance cues. As observed in \Cref{fig:pseudomask} and prior work~\cite{kulhanek2024wildgaussians}, raw MSE is not very informative when renderings are imperfect; instead, we aim for a less noisy pseudo-mask than the coarse error maps used in optimization-based pipelines.

We first extract DINOv3 patch embeddings and define a semantic dissimilarity map \(D_{\text{DINO}}(p) = 1 - \langle \Phi_p(I), \Phi_p(\hat{I}) \rangle\), where \(\Phi_p(\cdot)\) are L2-normalized patch features and \(\langle \cdot,\cdot\rangle\) is the inner product. In parallel, we compute a pixel-level appearance dissimilarity via SSIM, \(D_{\text{SSIM}}(x) = 1 - \mathrm{SSIM}(I,\hat{I})(x)\). For sharp motion segmentation, we operate at patch resolution: \(D_{\text{SSIM}}\) is downsampled to the patch grid by area pooling, and both maps are z-score normalized, \(\mathcal{Z}(D) = (D - \mu_D)/\sigma_D\).

We then fuse the two dissimilarities with \emph{adaptive} weights that depend on the current rendering fidelity. Let \(w_{\text{DINO}}, w_{\text{SSIM}} \in [0,1]\) with \(w_{\text{DINO}} + w_{\text{SSIM}} = 1\). When renderings are still coarse, we upweight the more stable semantic cue from DINOv3 (larger \(w_{\text{DINO}}\)); as training progresses and photometric quality improves, we gradually increase \(w_{\text{SSIM}}\) to exploit fine-grained appearance differences. The fused saliency is
$
D_{\text{bin}}(p) = w_{\text{DINO}} \,\mathcal{Z}(D_{\text{DINO}}(p)) + w_{\text{SSIM}} \,\mathcal{Z}(D_{\text{SSIM}}(p)).
$

We cluster all patch embeddings \(\{\Phi_p(I)\}\) across a batch of $B$ frames from the same scene using K-means to group semantically similar regions. A cluster $k$ is marked as foreground (motion) if its mean saliency \(\bar{s}_k = \mathbb{E}_{p \in k}[D_{\text{bin}}(p)]\) lies in the top $5\%$ and it is salient (above the $75$th percentile of \(D_{\text{bin}}\)) in at least $4$ frames, enforcing cross-frame consistency. Selected clusters are upsampled to pixel resolution via nearest-neighbor interpolation and refined with (a) morphological smoothing, (b) small connected-component removal, and (c) GrabCut~\cite{rother2004grabcut} boundary refinement seeded by eroded foreground masks. This yields a clean binary mask \(M_{\text{bin}} \in \{0,1\}^{H \times W}\) with sharp boundaries.

All clustering and fusion are performed at patch resolution, reducing computation by roughly $100\times$ while preserving motion boundaries. This is crucial in our learning-based setting, where we cannot afford to precompute and store DINOv3 features for the entire dataset.

\newcommand{\best}[1]{\textbf{#1}}
\newcommand{\second}[1]{\underline{#1}}
\definecolor{tablered}{RGB}{255,200,200}      % best
\definecolor{tableorange}{RGB}{255,230,180}   % second best
\definecolor{tableyellow}{RGB}{255,255,200}   % third best
% -------------------------------------------
\begin{table*}[t]
\centering
\scriptsize
\setlength{\tabcolsep}{1.8pt}
\renewcommand{\arraystretch}{1.12}
\resizebox{\linewidth}{!}{
% \begin{tabular}{l *{18}{c}}
\newcolumntype{Y}{>{\centering\arraybackslash}p{9mm}}
\begin{tabular}{l *{18}{Y}}
\toprule
& \multicolumn{9}{c}{\textbf{D-RE10K}} & \multicolumn{9}{c}{\textbf{D-RE10K-iPhone}} \\
\cmidrule(lr){2-10}\cmidrule(lr){11-19}
& \multicolumn{3}{c}{\textbf{Views = 2}} & \multicolumn{3}{c}{\textbf{Views = 3}} & \multicolumn{3}{c}{\textbf{Views = 4}}
& \multicolumn{3}{c}{\textbf{Views = 2}} & \multicolumn{3}{c}{\textbf{Views = 3}} & \multicolumn{3}{c}{\textbf{Views = 4}} \\
\cmidrule(lr){2-4}\cmidrule(lr){5-7}\cmidrule(lr){8-10}
\cmidrule(lr){11-13}\cmidrule(lr){14-16}\cmidrule(lr){17-19}
\textbf{Method}
& PSNR$_s\!\uparrow$ & SSIM$_s\!\uparrow$ & LPIPS$_s\!\downarrow$
& PSNR$_s\!\uparrow$ & SSIM$_s\!\uparrow$ & LPIPS$_s\!\downarrow$
& PSNR$_s\!\uparrow$ & SSIM$_s\!\uparrow$ & LPIPS$_s\!\downarrow$
& PSNR$\!\uparrow$ & SSIM$\!\uparrow$ & LPIPS$\!\downarrow$
& PSNR$\!\uparrow$ & SSIM$\!\uparrow$ & LPIPS$\!\downarrow$
& PSNR$\!\uparrow$ & SSIM$\!\uparrow$ & LPIPS$\!\downarrow$ \\
\midrule
\multicolumn{19}{c}{\textit{Optimization-Based Methods}}\\
\midrule
NeRF On-the-go~\cite{ren2024nerf}
& 15.90 & 0.518 & 0.582 & 18.45 & 0.624 & 0.446 & 19.52 & 0.620 & 0.443
& 17.04 & 0.412 & 0.636 & 17.53 & 0.484 & 0.581 & 17.04 & 0.470 & 0.616 \\
3DGS~\cite{kerbl20233d}
& 13.49 & 0.442 & 0.605 & 14.92 & 0.514 & 0.531 & 16.28 & 0.552 & 0.490
& 13.80 & 0.411 & 0.600 & 15.94 & 0.517 & 0.516 & 16.30 & 0.529 & 0.503 \\
T-3DGS~\cite{markin2024t}
& 15.90 & 0.518 & 0.582 & 18.45 & 0.624 & 0.446 & 18.70 & 0.613 & 0.454
& 14.07 & 0.380 & 0.632 & 15.61 & 0.428 & 0.573 & 16.95 & 0.525 & 0.487 \\
Spotless-Splats~\cite{sabour2025spotlesssplats}
& 16.45 & \cellcolor{tableyellow}0.548 & \cellcolor{tableyellow}0.468
& 17.77 & 0.600 & 0.394
& 18.05 & 0.608 & 0.390
& 17.73 & \cellcolor{tableorange}0.569 & \cellcolor{tableorange}0.424
& 18.65 & \cellcolor{tableorange}0.606 & \cellcolor{tableorange}0.385
& 18.86 & 0.615 & 0.382 \\
WildGaussians~\cite{kulhanek2024wildgaussians}
& 16.12 & 0.512 & 0.624 & 17.76 & 0.577 & 0.588 & 18.11 & 0.597 & 0.574
& \cellcolor{tableyellow}18.43 & 0.514 & 0.643
& \cellcolor{tableorange}19.82 & 0.542 & 0.637
& \cellcolor{tableorange}20.44 & 0.571 & 0.620 \\
\midrule
\multicolumn{19}{c}{\textit{Feed-forward Methods}}\\
\midrule
RayZer + Co-Seg~\cite{amir2021deep}
& \cellcolor{tableyellow}16.76 & 0.547 & 0.516 & 17.53 & 0.565 & 0.580 & 18.67 & 0.636 & 0.481
& 16.06 & 0.446 & 0.645 & 17.09 & 0.522 & 0.561 & 17.96 & 0.685 & 0.473 \\
RayZer + MegaSAM~\cite{li2025megasam}
& -- & -- & -- & \cellcolor{tableorange}20.13 & \cellcolor{tableorange}0.688 & \cellcolor{tableorange}0.336
& \cellcolor{tableorange}20.66 & \cellcolor{tableorange}0.702 & \cellcolor{tableyellow}0.310
& -- & -- & -- & 18.52 & 0.516 & 0.502 & \cellcolor{tableyellow}19.98 & \cellcolor{tableorange}0.718 & \cellcolor{tableyellow}0.372 \\
RayZer + SAV~\cite{huang2025segment}
& \cellcolor{tableorange}19.01 & \cellcolor{tableorange}0.628 & \cellcolor{tableorange}0.397
& \cellcolor{tableyellow}20.35 & \cellcolor{tableyellow}0.696 & \cellcolor{tableyellow}0.332
& \cellcolor{tableyellow}20.73 & \cellcolor{tableyellow}0.711 & \cellcolor{tableorange}0.308
& \cellcolor{tableorange}19.57 & \cellcolor{tableyellow}0.558 & \cellcolor{tableyellow}0.428
& \cellcolor{tableyellow}19.94 & \cellcolor{tableyellow}0.588 & \cellcolor{tableyellow}0.406
& \cellcolor{tableyellow}20.01 & \cellcolor{tableyellow}0.704 & \cellcolor{tableorange}0.355 \\
\textbf{WildRayZer (ours)}
& \cellcolor{tablered}21.78 & \cellcolor{tablered}0.734 & \cellcolor{tablered}0.308
& \cellcolor{tablered}21.98 & \cellcolor{tablered}0.754 & \cellcolor{tablered}0.314
& \cellcolor{tablered}22.38 & \cellcolor{tablered}0.773 & \cellcolor{tablered}0.290
& \cellcolor{tablered}20.89 & \cellcolor{tablered}0.611 & \cellcolor{tablered}0.364
& \cellcolor{tablered}20.91 & \cellcolor{tablered}0.633 & \cellcolor{tablered}0.354
& \cellcolor{tablered}20.98 & \cellcolor{tablered}0.734 & \cellcolor{tablered}0.298 \\
\bottomrule
\end{tabular}}
\caption{\textbf{Main Results on Novel View Synthesis.}
We report mean performance for 2, 3, 4 input views on D-RE10K (left, static regions only)
and D-RE10K-iPhone (right, full-image fidelity).
Metrics are \(\mathrm{PSNR}\!\uparrow\), \(\mathrm{SSIM}\!\uparrow\), and \(\mathrm{LPIPS}\!\downarrow\).
Cells highlighted in \cellcolor{tablered}{\strut red}, \cellcolor{tableorange}{\strut orange}, and \cellcolor{tableyellow}{\strut yellow}
denote the best, second, and third results respectively. SAV denotes Segment Any Motion in Videos~\cite{huang2025segment}.}
\vspace{-1mm}
\label{tab:nvs_main}
\end{table*}

\paragraph{Masking Dynamic Input Tokens.}
To prevent transient content from entering the scene representation, we mask input tokens that correspond to predicted dynamic regions. For each input view, the motion estimator produces a probability map, which we downsample to the token grid to obtain a patch-level motion score. Patches whose score exceeds a threshold are marked as dynamic, and their fused tokens are simply zeroed out before being fed into the scene encoder $\mathcal{E}_{\mathrm{scene}}$. Only static tokens participate in scene reconstruction, ensuring that moving objects do not leak into the static scene tokens.

\vspace{-2mm}
\paragraph{Copy--Paste Augmentation.}
To further improve the robustness of motion estimation and masking, we inject synthetic transients using a simple copy--paste strategy~\cite{ghiasi2021simple,wang2024videocutler}. We randomly sample objects from COCO~\cite{lin2014microsoft}, apply their provided segmentation masks, and paste them onto training images at random scales and positions. These pasted regions are treated as ground-truth transient masks and override the model’s predicted motion scores for those pixels. This augmentation exposes the system to a broader distribution of moving objects, strengthens open-set generalization, and provides clean supervision for token removal without altering the masking mechanism itself. See out-of-domain visualizations in \Cref{fig:additional_vis}. We provide details in appendix.

\section{Experiments}
\label{sec:exp}
\begin{figure*}[t]
    \centering
    \includegraphics[width=\linewidth]{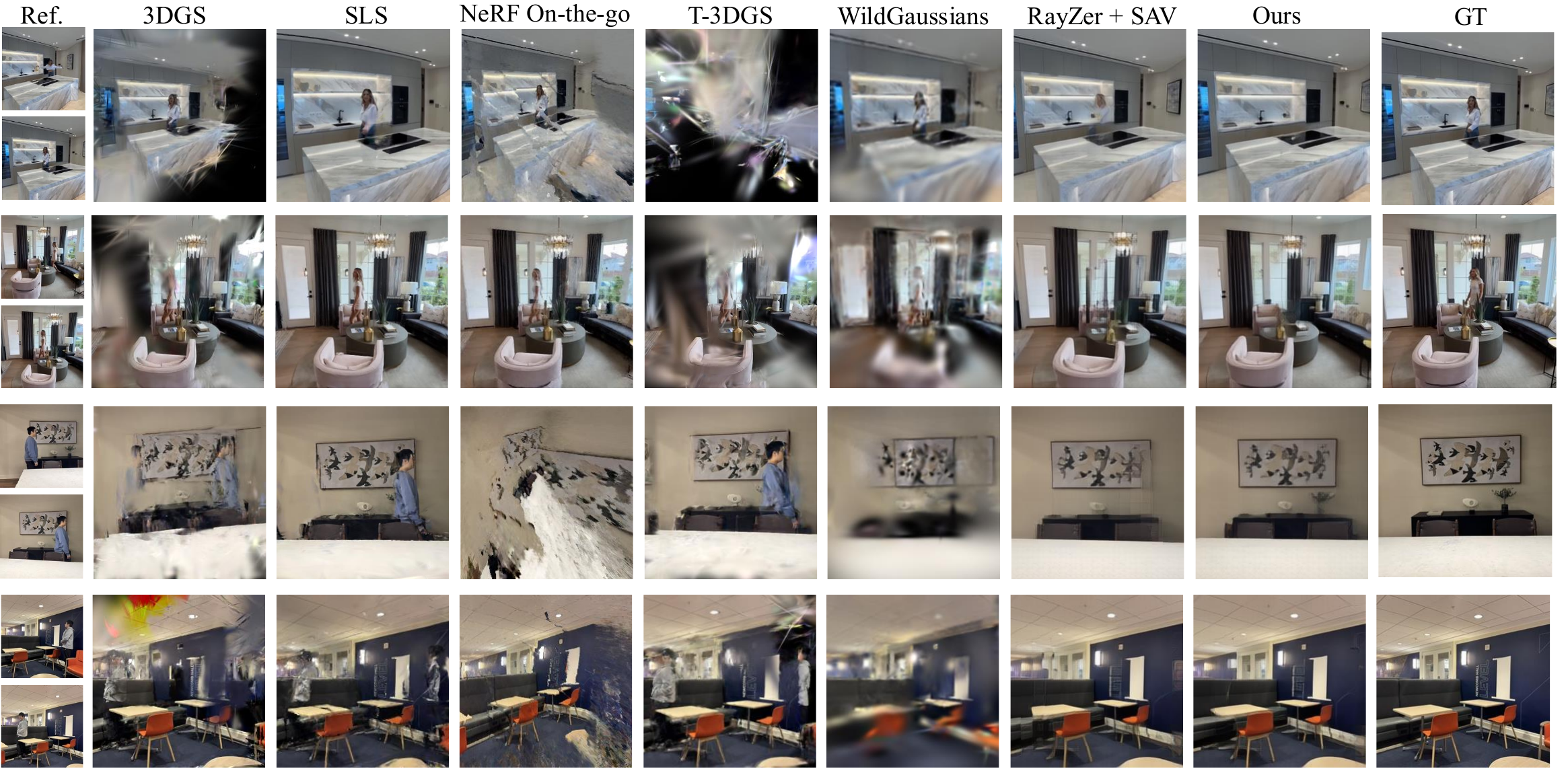}
        \caption{
            \textbf{Qualitative Comparisons.} Qualitative results on DRE10K-Mask (top two rows) and DRE10K-iPhone (bottom row). Compared to baselines, our method (1) more cleanly removes transient objects, (2) better handles cross-view completion (compare with RayZer + SAV baseline), and (3) better preserves global scene geometry (\eg, kitchens) and fine details (\eg, plants). SLS denotes Splotless-Splats~\cite{sabour2025spotlesssplats}.
        }
    % \vspace{-2mm}
    \label{fig:qualitative_main}
    \vspace{-2mm}
\end{figure*}

We now describe our experimental setup and present quantitative and qualitative results.  
WildRayZer uses 28 transformer layers: 4 for the motion estimator and 8 each for the camera estimator, scene encoder, and rendering decoder.  
Given our 4$\times$H100 compute budget, we train RayZer with $768$ scene tokens using a learning rate of $4\times 10^{-4}$, a cosine schedule for 100k iterations, and a batch size of 64.  
The perceptual-loss weight is $\lambda = 0.2$.  
For motion-mask training, we use a learning rate of $2\times 10^{-4}$, and $1\times 10^{-4}$ for training the masked renderer.  
All models operate at $256^2$ resolution with patch size 16. Additional implementation details are provided in the Appendix.

\subsection{Experiment Setup}

We describe the datasets, evaluation protocol, metrics, and baseline methods used in our experiments.

\vspace{-4mm}
\paragraph{Datasets.}
We evaluate on two challenging dynamic-scene benchmarks.  
D-RE10K-Mask contains 74 Internet-curated indoor sequences with moving humans, pets, and vehicles; each frame is paired with human-verified motion masks, enabling evaluation restricted to transient regions.  
D-RE10K-iPhone consists of 50 real-world sequences captured using a tripod-mounted iPhone.

\vspace{-4mm}
\paragraph{Evaluation Protocol and Metrics.}
We adopt a sparse-view NVS protocol: two, three or four input images are used for pose and scene estimation, and six target views are used for evaluation.  
We report static-region metrics for D-RE10K-Mask and report full-image metrics for D-RE10K-iPhone.
Image quality is evaluated using PSNR~\cite{psnr}, SSIM~\cite{wang2004image}, and LPIPS~\cite{zhang2018perceptual}.  
Furthermore, when evaluating motion mask, motion-mask accuracy is measured using mIoU and Recall against human verified annotations. 
We provide details about evaluation protocol in appendix.

\vspace{-4mm}
\paragraph{Baselines.}
We compare against state-of-the-art dynamic-scene NVS methods, including NeRF On-the-go~\cite{ren2024nerf}, 3DGS~\cite{kerbl20233d}, T-3DGS~\cite{markin2024t}, WildGaussians~\cite{kulhanek2024wildgaussians}, and Spotless-Splats~\cite{sabour2025spotlesssplats}.  
All baselines are re-evaluated under the same sparse-view setting.  
WildGS-SLAM~\cite{zheng2025wildgs} is excluded because its renderer is not public and its SLAM assumptions are incompatible with our few-view regime.

We further implement three RayZer-based variants using off-the-shelf motion estimators:  
(a) training-free co-segmentation~\cite{amir2021deep},  
(b) Segment Any Motion (SAV)~\cite{huang2025segment}, and  
(c) MegaSAM~\cite{li2025megasam}.  
Their predicted masks are downsampled to the token grid and used to \emph{drop} corresponding image tokens before scene aggregation.  
All variants are tuned on the D-RE10K validation set and evaluated without modification on the iPhone benchmark.

\subsection{Main Results}
As shown in \Cref{tab:nvs_main} and \Cref{fig:qualitative_main}, WildRayZer consistently outperforms all baselines.
Optimization-based pipelines struggle to suppress transients and reconstruct 3D scene structure from only 2–4 views. WildGaussians is performing better, but its rendering fidelity drops significantly under sparse inputs.  
In contrast, WildRayZer is feed-forward, pose-free at test time, and produces sharper background reconstructions with reliable transient removal.  
On D-RE10K-iPhone, our model also best recovers occluded background revealed across views, yielding the strongest full-image results.  
A detailed static vs.\ transient breakdown is provided in \Cref{subsec:analysis_rendering}.

Among RayZer-based variants, co-segmentation~\cite{amir2021deep} often over-masks under sparse viewpoints, leading to severe artifacts; we tune its hyper-parameters on the D-RE10K validation set. MegaSAM~\cite{li2025megasam} provides useful cues, but its boundaries are diffuse and predictions remain noisy with limited inputs. SAV~\cite{huang2025segment} relies on a pretrained tracker to select objects and can lock onto the wrong target under sparse-view settings. Finally, even with accurate off-the-shelf masks, naively \emph{masking} tokens yields blurry artifacts (akin to local interpolation), as visualized by the RayZer+SAV baseline in \Cref{fig:qualitative_main}. This suggests that cross-view completion requires learning, rather than masking alone.

\begin{figure*}[t]
    \centering
    \includegraphics[width=\linewidth]{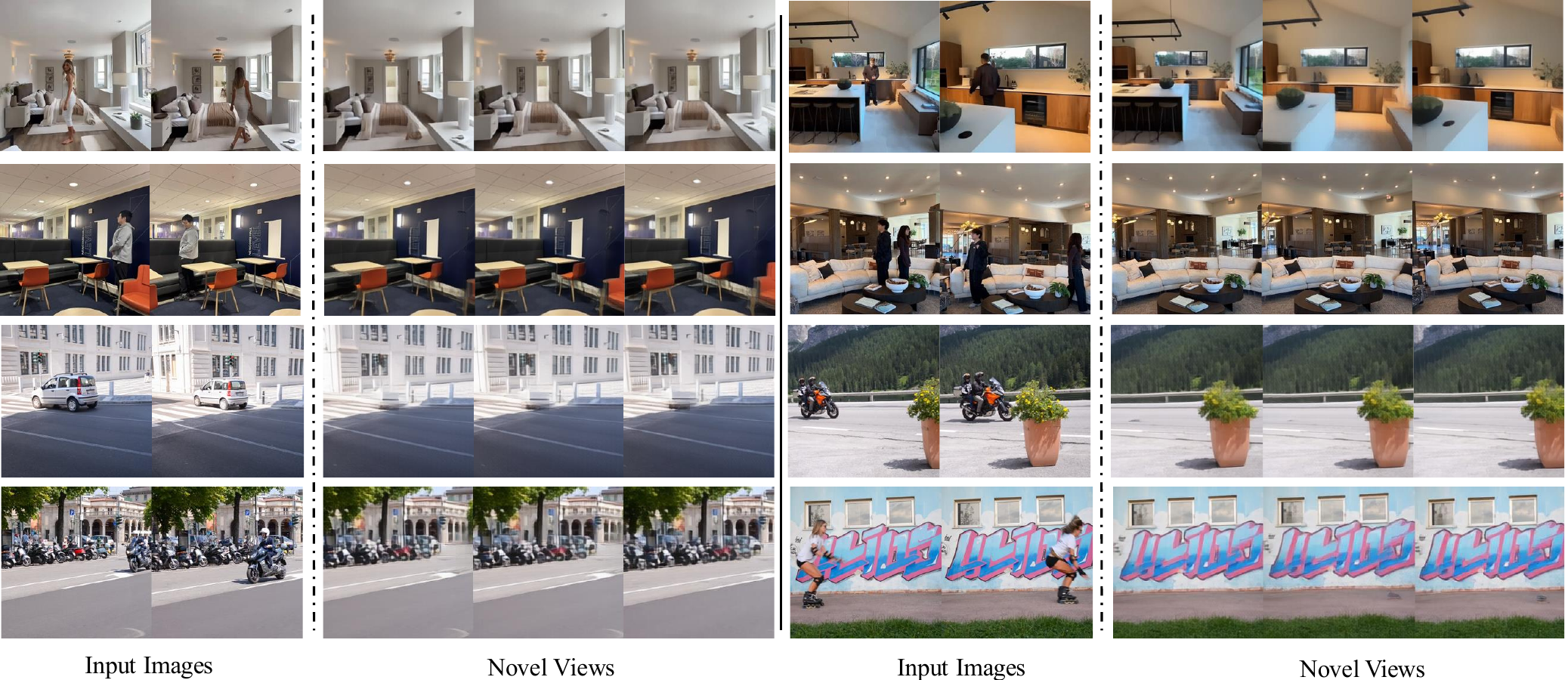}
    \vspace{-5mm}
    \caption{
        \textbf{Qualitative Results.} (1) First row: D-RE10K (no ground-truth novel views). (2) Second row: D-RE10K-iPhone. (3) Third and fourth rows: additional NVS results on DAVIS~\cite{perazzi2016benchmark}, where ground truth is also unavailable, demonstrating that WildRayZer generalizes to outdoor scenes and can mask unseen transient objects.
    }
    \label{fig:additional_vis}
    \vspace{-2mm}
\end{figure*}
\subsection{Analysis}

\paragraph{Analysis of Rendering.}
\label{subsec:analysis_rendering}
\definecolor{tablered}{RGB}{255,200,200}      % best
\definecolor{tableorange}{RGB}{255,230,180}   % second best
\definecolor{tableyellow}{RGB}{255,255,200}   % third best
% -------------------------------------------

\begin{table}[t]
\centering
\scriptsize
\setlength{\tabcolsep}{2.5pt}
\renewcommand{\arraystretch}{1.12}
\resizebox{\linewidth}{!}{
\begin{tabular}{l *{6}{c}}
\toprule
\textbf{Method}
& PSNR$_s\!\uparrow$ & SSIM$_s\!\uparrow$ & LPIPS$_s\!\downarrow$
& PSNR$_t\!\uparrow$ & SSIM$_t\!\uparrow$ & LPIPS$_t\!\downarrow$ \\
\midrule
\multicolumn{7}{c}{\textit{Optimization-Based Methods}}\\
\midrule
NeRF On-the-go~\cite{ren2024nerf}
& 17.11 & 0.419 & 0.633 & 16.98 & 0.323 & 0.667 \\
3DGS~\cite{kerbl20233d}
& 13.85 & 0.417 & 0.596 & 14.63 & 0.337 & 0.615\\
T-3DGS~\cite{markin2024t}
& 14.42 & 0.387 & 0.630 & 13.98 & 0.292 & 0.662 \\
Spotless-Splats~\cite{sabour2025spotlesssplats}
& 17.91 & \cellcolor{tableyellow}0.575 & \cellcolor{tableyellow}0.420
& 17.42 & \cellcolor{tableorange}0.494 & \cellcolor{tableorange}0.477 \\
WildGaussians~\cite{kulhanek2024wildgaussians}
& 18.47 & 0.518 & 0.642
& \cellcolor{tableorange}18.46 & \cellcolor{tableyellow}0.449 & 0.652 \\
\midrule
\multicolumn{7}{c}{\textit{Feed-forward Methods}}\\
\midrule
RayZer + Co-Seg~\cite{amir2021deep}
& \cellcolor{tableyellow}20.05 & 0.565 & 0.469 & 15.74 & 0.363 & 0.621 \\
RayZer + SAV~\cite{huang2025segment}
& \cellcolor{tableorange}20.22 & \cellcolor{tableorange}0.579 & \cellcolor{tableorange}0.406
& \cellcolor{tableyellow}17.77 & 0.437 & \cellcolor{tableyellow}0.536 \\
\textbf{WildRayZer (ours)}
& \cellcolor{tablered}21.00 & \cellcolor{tablered}0.612 & \cellcolor{tablered}0.360
& \cellcolor{tablered}20.99 & \cellcolor{tablered}0.575 & \cellcolor{tablered}0.371 \\
\bottomrule
\end{tabular}
}
\caption{\textbf{Novel View Synthesis on \textbf{D-RE10K-iPhone} (Views=2).}
Metrics are split into static (s) and transient (t) regions: \(\mathrm{PSNR}_s\!\uparrow\), \(\mathrm{SSIM}_s\!\uparrow\), \(\mathrm{LPIPS}_s\!\downarrow\), and \(\mathrm{PSNR}_t\!\uparrow\), \(\mathrm{SSIM}_t\!\uparrow\), \(\mathrm{LPIPS}_t\!\downarrow\). \vspace{-2mm}
}
\label{tab:rendering_analysis}
\vspace{-1mm}
\end{table}
To better analyze differences between methods, we separately evaluate static and transient regions on D-RE10K-iPhone with two input views ($v{=}2$; see Table~\ref{tab:rendering_analysis}).
Because transient objects typically cover only small image areas, full-image PSNR can be misleading.
Optimization-based methods show large gaps between static and transient metrics, while WildGaussians~\cite{kulhanek2024wildgaussians} is the only baseline that achieves relatively balanced performance in the $v{=}2$ setting, but with lower overall image quality.
Feed-forward baselines without a learned masked renderer exhibit strong occlusion artifacts, indicating that cross-view completion needs to be learned rather than purely inferred from static geometry.
WildRayZer improves performance in both static and transient regions by explicitly modeling motion masks and maintaining robust pose estimation under sparse, dynamic inputs.

\paragraph{Analysis of Motion Masks.}
\begin{table}[t]
\centering
\resizebox{\columnwidth}{!}{%
\begin{tabular}{lcccccc}
\toprule
& \multicolumn{2}{c}{$n{=}2$} & \multicolumn{2}{c}{$n{=}3$} & \multicolumn{2}{c}{$n{=}8$} \\
\cmidrule(lr){2-3}\cmidrule(lr){4-5}\cmidrule(lr){6-7}
\textbf{Method} & mIoU $\uparrow$ & Recall $\uparrow$ & mIoU $\uparrow$ & Recall $\uparrow$ & mIoU $\uparrow$ & Recall $\uparrow$ \\
\midrule
\multicolumn{7}{c}{\textit{Supervised Methods}}\\
\midrule
MegaSAM~\cite{li2025megasam}                &  -- &  -- & 12.7 & 37.1 & 35.4 & 60.7 \\
Segment Any Motion~\cite{huang2025segment} &  31.9  &  47.2   &  41.2   &  57.1   &  50.9   &  70.8   \\
\midrule
\multicolumn{7}{c}{\textit{Self-supervised Methods}}\\
\midrule
Co-segmentation~\cite{amir2021deep}        &  9.6 & 45.0 & 13.7 & 53.2 & 16.3 & 45.5 \\
WildRayZer             &  \textbf{53.9}  &  \textbf{85.1}   &  \textbf{52.1}   &  \textbf{84.3}   &  \textbf{54.2}   &  \textbf{87.7}   \\
\bottomrule
\end{tabular}%
}
\caption{\textbf{Motion-mask quality.} Comparison of supervised and self-supervised motion segmentation methods under sparse-view settings. WildRayZer achieves higher mIoU and recall across different numbers of input views.}
\label{tab:dre10k_motion_mask_full}
\vspace{-2mm}
\end{table}

We further evaluate motion-mask quality across methods.  
Co-segmentation with DINOv2~\cite{oquab2023dinov2} over-segments, hurting NVS quality.  
MegaSAM is sensitive to thresholding and becomes noisy in sparse-view setting.  
SAV~\cite{wang2024videocutler} often chooses irrelevant objects under sparse views.  
All baseline models improve with more views.  
As an oracle check, we run Segment-Any-Motion on 25-frame segments and observe strong agreement with our GT annotations (mIoU $>$ 65).  
Additional examples are shown in the appendix.

\subsection{Ablation Study}

\paragraph{Effects of Copy–Paste Augmentation.}
\begin{table}[t]
\centering
\footnotesize
\setlength{\tabcolsep}{4.5pt}
\renewcommand{\arraystretch}{0.95}
\begin{tabular}{lcc}
\toprule
Training Recipe & D-RE10K & D-RE10K-iPhone \\
\midrule
Copy–Paste Only            & 18.2 & 11.1 \\
Pseudo–Mask Only           & 53.9 & 45.3 \\
Copy–Paste + Pseudo–Mask   & \textbf{53.9} & \textbf{49.7} \\
\bottomrule
\end{tabular}
\caption{\textbf{Copy–Paste Ablation.} Copy–paste alone does not transfer to real videos but improves out-of-domain generalization when combined with pseudo-masks.    \vspace{-4mm}}
\label{tab:mask_evolution}
\end{table}
We isolate copy–paste augmentation by freezing the renderer and training only the motion estimator.  
Copy–paste alone does not produce meaningful masks, but when added on top of D-RE10K pretraining, it improves cross-dataset generalization.  
The motion estimator transfers to DAVIS~\cite{perazzi2016benchmark}, where copy–paste improves mIoU from 3.4 to 31.0 on a subset of eight sequences with clear motion and sufficient parallax. We provide visualizations on DAVIS~\cite{perazzi2016benchmark} in \Cref{fig:additional_vis}.

\vspace{-4mm}
\paragraph{Effects of Motion Estimator Input Modalities.}
Using both image tokens and camera-ray tokens allows gradients to flow into the pose estimator, leading to more stable camera estimation in dynamic scenes.
Incorporating DINOv3 features substantially accelerates mask emergence and improves their quality: reaching $\mathrm{mIoU}{=}30$ takes about $20\text{k}$ steps without DINOv3 but only $1.5\text{k}$ with it.
Final performance also increases (39.4 vs.\ 29.4 mIoU on D-RE10K), which is particularly important given the cost of each pseudo-mask iteration.

\section{Conclusions}
\label{sec:conclusion}

In this work, we introduced WildRayZer, a feed-forward SSL framework for novel view synthesis in dynamic environments that disentangles object motion from static 3D structure. Across sparse-view, transient-aware NVS and motion segmentation benchmarks, WildRayZer consistently outperforms each baseline. We further curated D-RE10K and D-RE10K-iPhone, providing large-scale training data and a benchmark for dynamic NVS under sparse views. Our results indicate that disentangling camera motion and object motion is feasible in a fully self-supervised setting; limitations and future directions are discussed in the supplementary material.

\section{Acknowledgment}

The authors acknowledge the Adobe Research Gift, the University of Virginia Research Computing and Data Analytics Center, Advanced Micro Devices AI and HPC Cluster Program, Advanced Cyberinfrastructure Coordination Ecosystem: Services \& Support (ACCESS) program, and National Artificial Intelligence Research Resource (NAIRR) Pilot for computational resources, including the Anvil supercomputer (National Science Foundation award OAC 2005632) at Purdue University and the Delta and DeltaAI advanced computing resources (National Science Foundation award OAC 2005572). The authors thank Wen Ying, Zhenyu Li and Jin Yao for constructing D-RE10K-iPhone, and thank Jiayun Wang, Hanwen Jiang, and Hao Tan for proofreading.
{
    \small
    \bibliographystyle{ieeenat_fullname}
    \bibliography{main}
}

\clearpage
\appendix
\setcounter{page}{1}
\maketitlesupplementary

\Cref{sec:training-details} details the full WildRayZer training pipeline, including RayZer pretraining, motion-mask learning, masked reconstruction, and joint training with copy–paste augmentation.
\Cref{sec:benchmarking} details the D-RE10K-iPhone benchmark and clarifies evaluation metrics.
\Cref{sec:additional_vis} presents extended qualitative visualizations, including motion-mask comparisons, failure analyses, and additional results on D-RE10K, D-RE10K-iPhone, and DAVIS~\cite{pont20172017}.
\section{Training Details}
\label{sec:training-details}

\paragraph{RayZer Pretraining.}
We first train a RayZer~\cite{jiang2025rayzer} using official release code with scene and pose latent representations. The encoder consists of 12 transformer layers with 8 additional geometry-specific layers, while the decoder uses 12 layers. We use a hidden dimension of $d=768$ with attention heads of size 64. Both encoder and decoder employ QK normalization for training stability. Images are tokenized into $16\times16$ patches from $256\times256$ input resolution. The scene latent code has length 768, and camera poses are represented using 6D rotation representation. Importantly, we find smaller scene latent code yields better rendering quality compared to scene latent code length of $3084$ when training with smaller batch size. Additionally, instead of relying solely on absolute positional embeddings, we introduce a dedicated embedding to distinguish input and target views. We also mix cases with 2, 3, and 4 input views during training so that RayZer can handle variable-length inputs without performance degradation.

We train on the RealEstate10K~\cite{zhou2018stereo} dataset with 2 input views and 6 target views per scene. The model is trained for 100K steps with a batch size of 8 per GPU and gradient accumulation over 2 steps. We use the AdamW~\cite{loshchilov2017decoupled} optimizer with $\beta_1=0.9$, $\beta_2=0.95$, learning rate of $4\times10^{-4}$, and weight decay of 0.05. The learning rate is warmed up over 6K steps. Gradient clipping is applied with a maximum norm of 1.0. We use mixed precision training (bfloat16) with TF32 enabled on H100.

The training objective combines L2 reconstruction loss (weight 1.0) and perceptual loss (weight 0.2) computed using VGG features. Input views are sampled with frame distances between 25 and 192 frames, with a scene scale factor of 1.35 applied during data augmentation.

\vspace{-3mm}
\paragraph{Motion Mask Training Stage.}
In this stage, we train the motion mask predictor while keeping the pre-trained rendering model frozen. The goal is to learn to distinguish dynamic regions from static using self-supervised pseudo-labels derived from DINOv3~\cite{simeoni2025dinov3} features and SSIM~\cite{wang2004image}.

We generate motion pseudo-labels by computing ground-truth and renderings differences semantically and structurally. For each pair of ground truth and rendering, we extract dense feature maps and compute cosine similarity between corresponding spatial locations. Then we derive a fused error map from SSIM dissimilarity map and DINO dissimilarity map, viewing as saliency map. Then we cluster DINO feature at patch level across input images and assign average saliency score based on the saliency map. Regions with low feature consistency across frames (similarity below a learned threshold) are labeled as dynamic, while consistent regions are labeled as static. We apply morphological operations (kernel size 3, 1 iteration) to expand the mask and filter out small components (area $< 0.25\%$ of image) to ensure coherent motion segments. Lastly, we use grabcut~\cite{tang2013grabcut} to refine the boundary.

The motion mask predictor is trained on D-RE10K using binary cross-entropy loss between predicted masks and pseudo motion labels. We use a learning rate of $10^{-4}$ with AdamW optimizer ($\beta_1=0.9$, $\beta_2=0.95$, weight decay 0.01) and batch size 32. Critically, we apply a PSNR-based sample filtering strategy: only samples with rendering PSNR $> 17$ dB are used for mask supervision. This prevents our model to learn from noisy pseudo-labels on challenging dynamic scenes where the frozen renderer struggles. We show detailed process of pseudo motion mask construction in \cref{fig:pseudomask}.

\vspace{-3mm}
\paragraph{Masked Latent Scene Reconstruction Stage.}
In this stage, we freeze the learned motion mask predictor and train the rendering model to reconstruct static scene content while explicitly providing masks on input images. This stage teaches the model to perform cross-view completion over masked areas.
We implement a mask token strategy during training by masking 10\% token, similar to standard MAE training but tailored for novel view synthesis. More specifically, we find clustered mask performances better compare to random token masking.
The reconstruction loss is computed on the entire image.

\vspace{-3mm}
\paragraph{Joint training with Copy-Paste Augmentation.}
In the final stage, we jointly train the motion-mask predictor and the rendering model with synthetic motion augmentation. Specifically, we adopt copy–paste augmentation~\cite{ghiasi2021simple} to insert controlled transient objects into static RE10K scenes, providing explicit supervision for the motion-mask predictor while simultaneously training the renderer to remain robust to these diverse transients. In addition, we mix D-RE10K sequences by masking the predicted motion regions, iteratively refining pseudo motion labels and excluding dynamic areas from the reconstruction loss.

\begin{figure}[t]
    \centering
    \includegraphics[width=\linewidth]{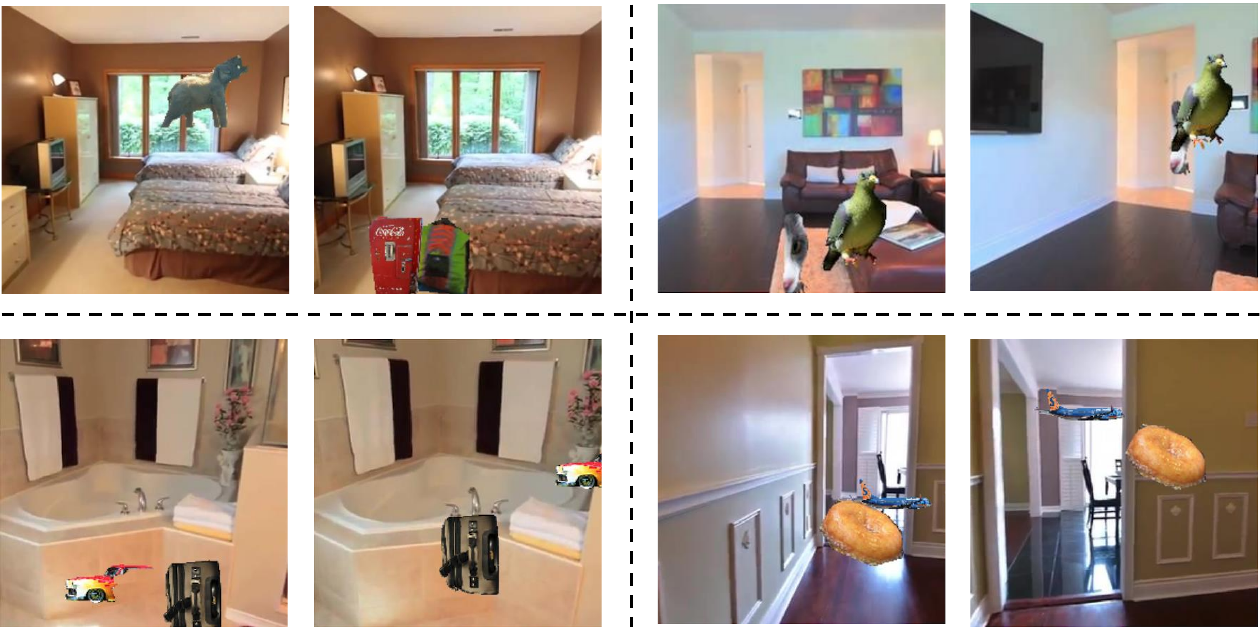}
    \vspace{-4mm}
    \caption{
        \textbf{Examples of Copy–paste mask augmentation.}
        We inject synthetic transient objects (\eg, animals, household items, vehicles) into static RE10K scenes
        to simulate dynamic elements in otherwise static environments. \vspace{-3mm}
    }
    \label{fig:copy-paste}
    \vspace{-2mm}
\end{figure}

We detail the copy-paste augmentation here and provide examples in Fig.~\ref{fig:copy-paste}. We randomly paste 1-2 COCO objects (animals, vehicles, people) into 50\% of training scenes occupying 25-35\% of image size. Objects are blended using Gaussian smoothing ($\sigma=3$) and positioned with 15\% margin from image borders. With 80\% probability, we use per-view random objects; otherwise, we paste the same object across all views in a sequence. This generates binary overlay masks $M_{\text{paste}}$ indicating pasted regions.

In this stage, the motion mask predictor receives supervision from two sources: (1) DINOv3 pseudo-labels $M_{\text{DINO}}$ for real dynamic content in D-RE10K, and (2) ground-truth paste masks $M_{\text{paste}}$ for synthetic objects in augmented Static RE10K. We use binary cross-entropy loss with PSNR filtering (threshold 17 dB) to ensure supervision quality. The rendering loss combines masked reconstruction on both static regions and pasted regions:
\begin{equation}
\mathcal{L} = \mathcal{L}_{\text{masked}} + \lambda_{\text{mask}} \cdot \text{BCE}(M_{\text{pred}}, M_{\text{target}}),
\end{equation}
where $M_{\text{target}} = M_{\text{DINO}}$ for dynamic scenes and $M_{\text{target}} = M_{\text{paste}}$ for augmented static scenes, and $\lambda_{\text{mask}} = 1.0$.

If we train only in the final stage, we observe many rendering artifacts due to inaccurate motion-mask predictions during early iterations. Therefore, adopting a progressive, multi-stage training strategy becomes essential for stability and reliable convergence.

\section{Benchmarking Details}
\label{sec:benchmarking}

In this section, we provide additional details of the proposed D-RE10K-iPhone benchmark in \Cref{subsec:detail-dre10k-iphone} and describe our masked image quality metrics in \Cref{subsec:detail-mask_metrics}. We will release our implementation as a reference.

\subsection{D-RE10K-iPhone Benchmark Details}
\label{subsec:detail-dre10k-iphone}
D-RE10K-iPhone comprises 50 real-world sequences featuring humans and vehicles.  
Each sequence contains 18 paired images captured from the same viewpoint: one \emph{transient} frame with a moving object (\eg, person, car) and one \emph{clean} frame without it.  
An iPhone mounted on a tripod with a Bluetooth shutter is used to minimize pose differences between paired images. Each sequence require roughly 30 minutes to curate.

To reduce illumination-induced artifacts, we compare static background regions across each pair and discard pairs with noticeable lighting changes (e.g., sudden occlusions or reappearance of direct sunlight).  
For the $v{=}2,3,4$ sparse-view settings, input views are selected to span the full range of available camera angles in each sequence, ensuring that we evaluate genuine novel view synthesis rather than near-view interpolation.

\subsection{Masked Image Quality Metrics}
\label{subsec:detail-mask_metrics}
We compute standard image quality metrics (PSNR~\cite{psnr}, SSIM~\cite{wang2004image}, LPIPS~\cite{zhang2018perceptual}) over spatial regions specified by real-valued masks $M \in [0,1]^{H\times W}$, enabling separate evaluation of transient and static scene components.

\noindent\textbf{Masked PSNR.}
Given a ground-truth image $I$, prediction $\hat{I}$, and mask $M$, we compute a masked mean squared error:
\begin{align}
\text{MSE}_M &=
\frac{\sum_{i,j} (I_{ij} - \hat{I}_{ij})^2 \, M_{ij}}{\sum_{i,j} M_{ij}}, \\[2pt]
\text{PSNR}_M &= -10 \log_{10}(\text{MSE}_M).
\end{align}
This yields a localized measure of pixel-level fidelity within the masked region.

\noindent\textbf{Masked SSIM.}
We first compute a standard SSIM map $S \in \mathbb{R}^{H'\times W'}$ using an $11\times 11$ Gaussian window ($\sigma = 1.5$) with stability constants $C_1 = (0.01)^2$ and $C_2 = (0.03)^2$ for images in $[0,1]$. Let $\mu$ and $\sigma$ denote local means and variances. The per-pixel SSIM is
\begin{equation}
S_{ij} = 
\frac{(2\mu_{ij}\hat{\mu}_{ij} + C_1)(2\sigma_{ij} + C_2)}
     {(\mu_{ij}^2 + \hat{\mu}_{ij}^2 + C_1)(\sigma_{ij}^2 + \hat{\sigma}_{ij}^2 + C_2)}.
\end{equation}
The mask is downsampled to the SSIM resolution via area pooling ($M'$), and the masked SSIM is
\begin{equation}
\text{SSIM}_M = 
\frac{\sum_{i,j} S_{ij}\, M'_{ij}}{\sum_{i,j} M'_{ij}}.
\end{equation}

\noindent\textbf{Masked LPIPS.}
To obtain perceptually weighted masked errors, we use LPIPS in spatial mode, producing per-layer feature difference maps $D^{(\ell)} \in \mathbb{R}^{H_\ell \times W_\ell}$. For each feature level $\ell$, we downsample the mask via area pooling to $M^{(\ell)}$ and compute
\begin{equation}
\text{LPIPS}_M = 
\sum_{\ell} 
\frac{\sum_{i,j} D^{(\ell)}_{ij}\, M^{(\ell)}_{ij}}
     {\sum_{i,j} M^{(\ell)}_{ij}}.
\end{equation}
This applies perceptual masking consistently across all LPIPS feature scales.

\begin{figure*}[t]
    \centering
    \includegraphics[width=\linewidth]{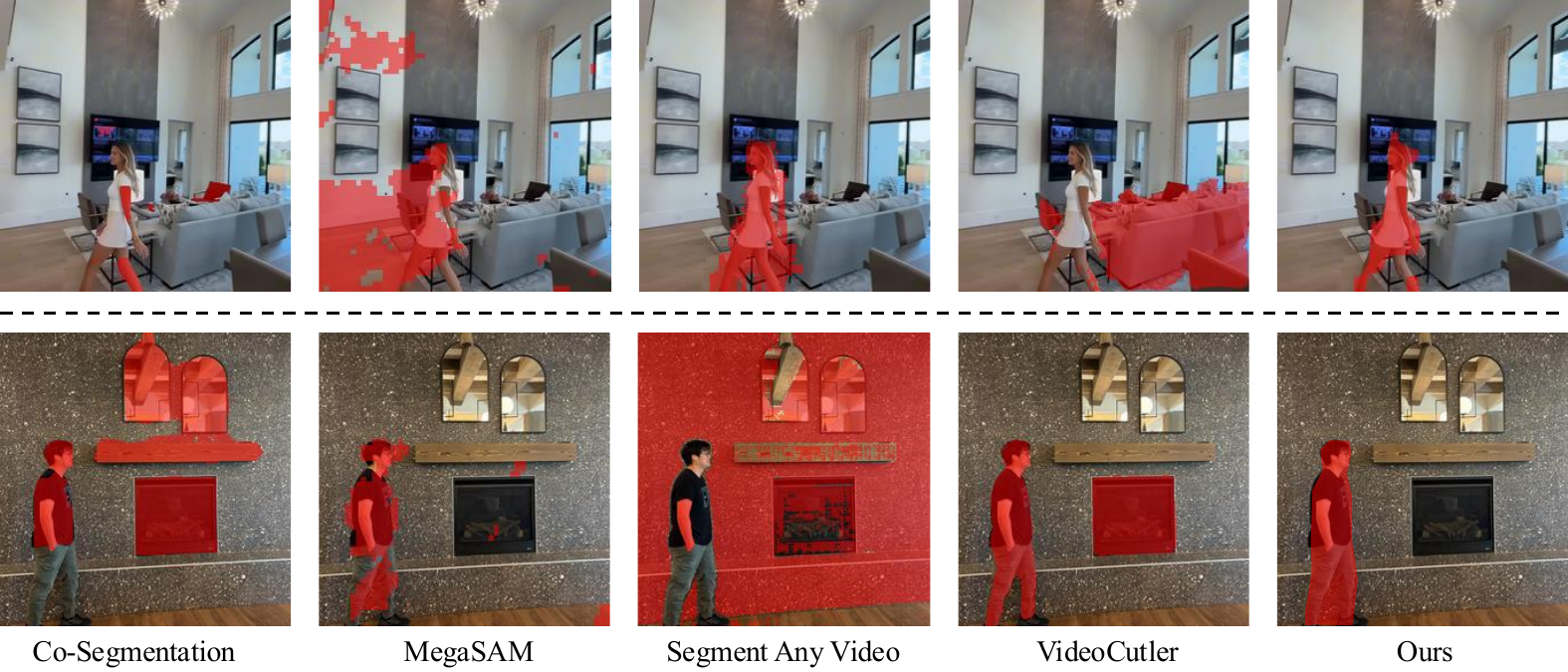}
    \vspace{-5mm}
    \caption{
        \textbf{Motion Masks Comparisons.}
        We present motion masks comparisons on D-RE10K (row 1) and D-RE10K-iPhone (row 2) across Co-segmentation~\cite{amir2021deep}, MegaSAM~\cite{li2025megasam}, Segment Any Video~\cite{huang2025segment}, VideoCutler~\cite{wang2024videocutler} and WildRayzer's motion mask predictions. \vspace{-3mm}
    }
    \label{fig:motion-mask}
    \vspace{-2mm}
\end{figure*}
\noindent\textbf{Implementation Details.}
All metrics are evaluated per image. For D-RE10K-iPhone's analysis, we report PSNR, SSIM, and LPIPS separately on transient regions (using $M_{\text{transient}}$) and on static regions (using $M_{\text{static}} = 1 - M_{\text{transient}}$). We notice different implementation can lead to different evaluation results, therefore we will release our implementations for these metrics as a reference.

\section{Additional Visualizations}
\label{sec:additional_vis}

In this section, we provide additional visualization: (1) motion mask comparisions with state-of-art methods in \Cref{subsec:motion_mask}; (2) present WildRayzer's failure cases in \Cref{figs/failure_cases}; (3) demonstrates additional WildRayzer's results on D-RE10K, D-RE10K-iPhone and DAVIS~\cite{pont20172017} benchmarks in \Cref{subsec:additional_vis_2}. 

\subsection{Motion Mask Visualization}
\label{subsec:motion_mask}
We compare several popular motion-mask generators: co-segmentation based on DINO features~\cite{amir2021deep}, MegaSAM~\cite{li2025megasam}, Segment Any Video (SAV)~\cite{huang2025segment}, and the self-supervised instance segmentation model VideoCutler~\cite{wang2024videocutler}. Even after careful hyperparameter tuning, training-free co-segmentation tends to either over-mask or under-mask large portions of the scene, which severely degrades rendering quality. MegaSAM provides useful cues, but under sparse-view inputs its masks are often noisy and inconsistent across views. SAV can be highly accurate when it selects the correct actor, but under sparse views it frequently segments the wrong object. VideoCutler, by design, returns salient instances rather than true movers, and thus does not distinguish dynamic from static objects; we visualize its first predicted mask for comparison. In contrast, our learned motion masks are the most faithful to true motion boundaries, despite being trained in a fully self-supervised manner without ground-truth motion labels. We demonstrate motion masks with one example from D-RE10K (top row) and D-RE10K-iPhone (bottom row) in Fig.~\ref{fig:motion-mask}.

\subsection{Failure Cases}
\label{figs/failure_cases}
\begin{figure*}[t]
    \centering
    \includegraphics[width=\linewidth]{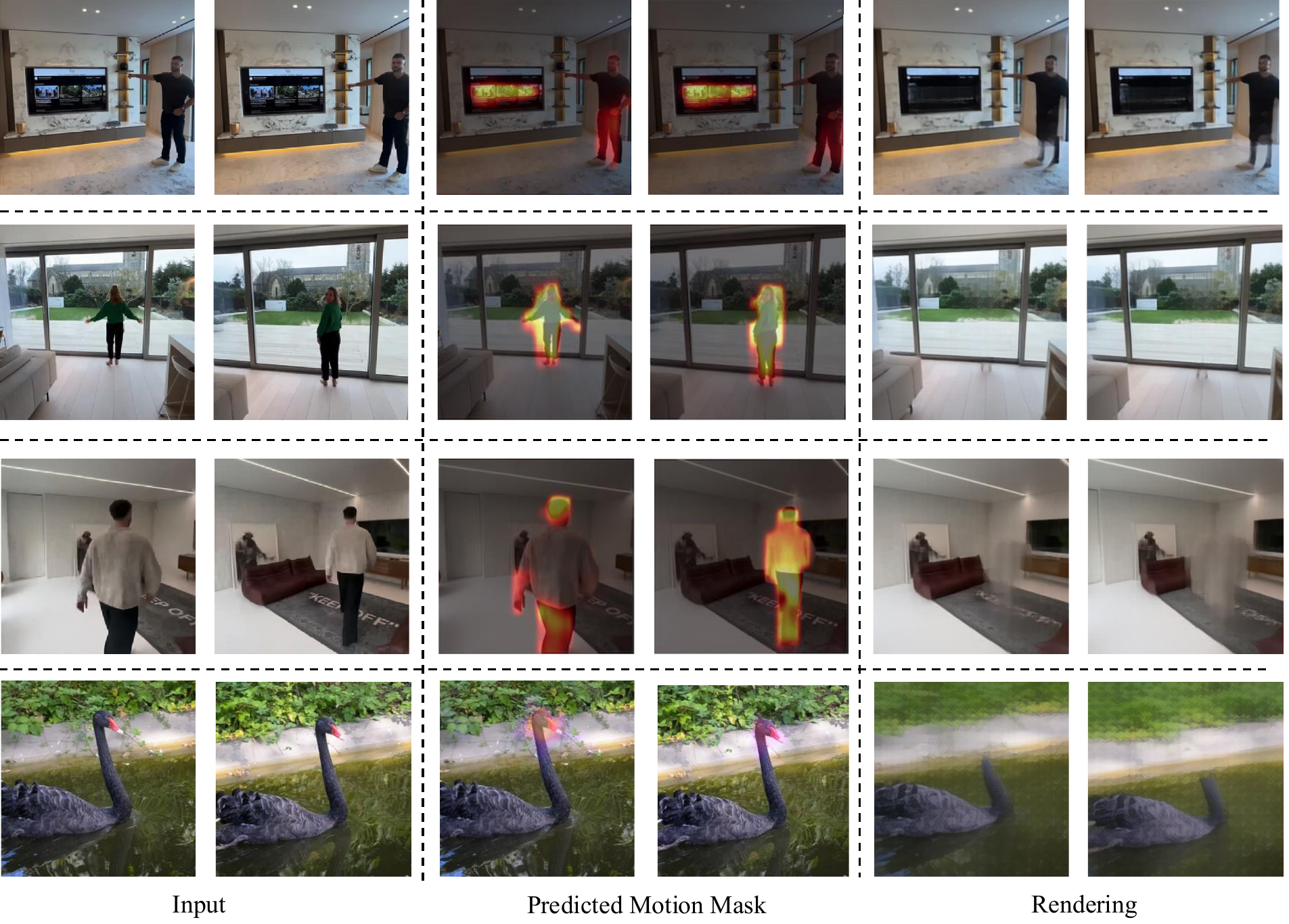}
    \vspace{-4mm}
    \caption{
        \textbf{Failure Cases.}
        Each row contains one failure case of WildRayzer. Motion mask only highlights moving  parts dispite highlight television, it only highlights part of human body in row  1. Moreover, motion mask may miss small places such as feet in row 2. Row 3 and 4 show failure case when transient object is too big  and predicted masks are smaller.
    }
    \label{fig:failure}
    \vspace{-2mm}
\end{figure*}

We illustrate common failure cases in \cref{fig:failure}. Because our pseudo motion masks do not enforce instance-level segmentation, they may capture only the moving parts of an object when other parts remain static across the input views. While this behavior can be consistent with the motion cue, it often degrades rendering quality and we treat it as a failure mode. We also observe under-segmentation, where the predicted motion mask is smaller than the true moving region (row 2). Finally, mask quality drops when the moving object occupies a large fraction of the input images (rows 3–4).
\subsection{Additional Visualizations}
\label{subsec:additional_vis_2}
\begin{figure*}[t]
    \centering
    \includegraphics[width=\linewidth]{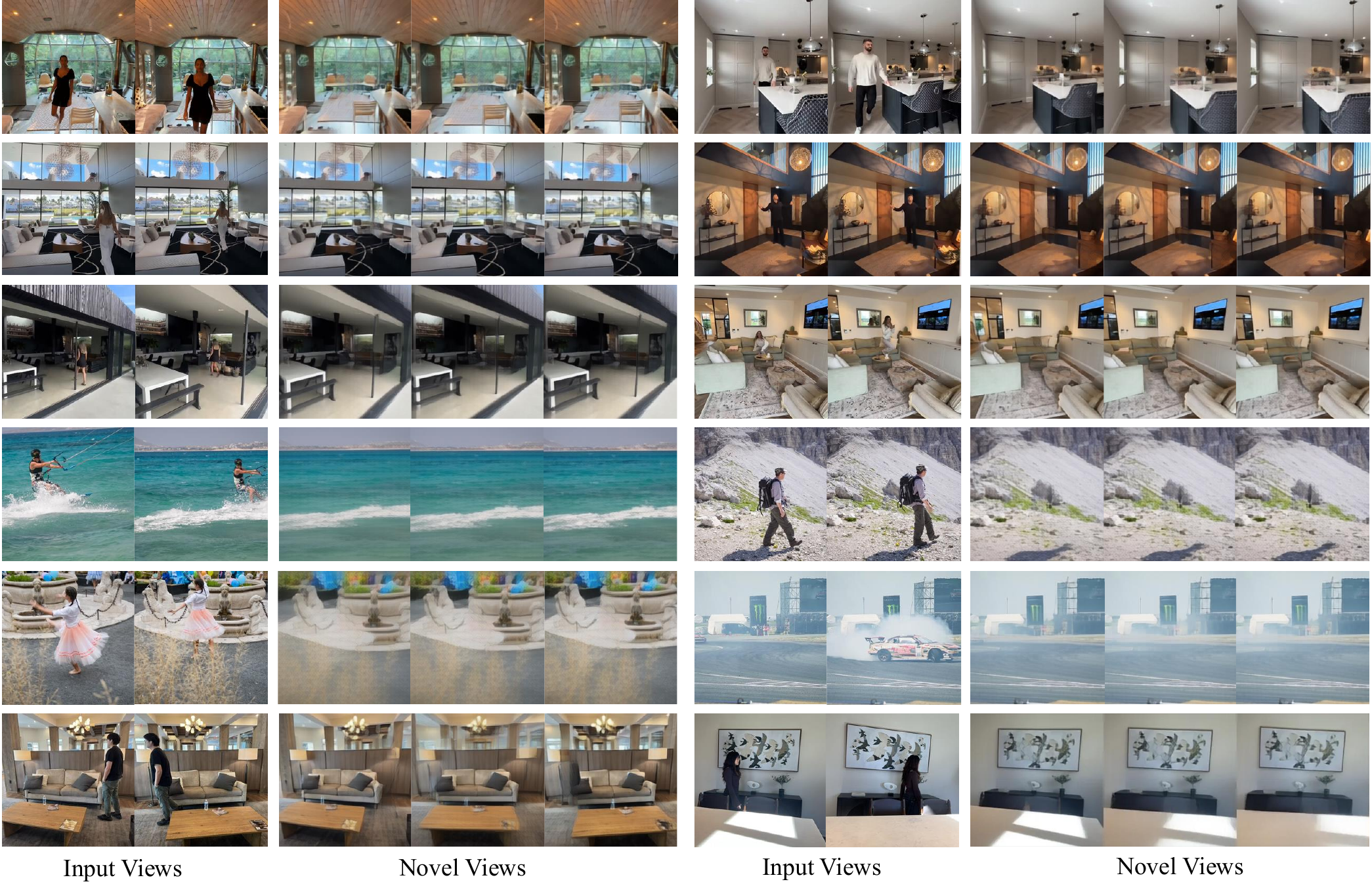}
    \vspace{-5mm}
    \caption{
    \textbf{Additional qualitative results.}
    We show 12 extra examples to illustrate WildRayZer’s behavior across datasets. 
    The first three rows are from D-RE10K, the next two rows demonstrate generalization to the unseen DAVIS dataset~\cite{pont20172017}, and the last row shows additional real-world results on D-RE10K-iPhone.
    }
    \label{fig:additional_qualitative}
    \vspace{-2mm}
\end{figure*}

We provide 12 additional qualitative examples to further illustrate the behavior of WildRayZer beyond the main paper results in Fig.~\ref{fig:additional_qualitative}. The first three rows depict challenging dynamic indoor scenes from D-RE10K; in these cases, WildRayZer successfully removes transients while plausibly complete occluded structure across input views. The next two rows demonstrate cross-dataset generalization on unseen DAVIS sequences~\cite{pont20172017}, showing that the learned motion masks and masked rendering transfer to videos with different content and capture conditions. The final row presents real-world examples from D-RE10K-iPhone under casual handheld capture, highlighting that our model maintains sharp static geometry and clean background reconstruction.

\end{document}